# Predicting Fuel Consumption in Power Generation Plants using Machine Learning and Neural Networks

Gabin Maxime Nguegnang
*African Institute for Mathematical Sciences (AIMS),* Limbe, Cameroon
maxime.nguegnang@aims-cameroon.org

Marcellin Atemkeng*
*Department of Mathematics*
Rhodes University, Grahamstown, South Africa, m.atemkeng@gmail.com

Theophilus Ansah-Narh
*Ghana Space Science & Technology Institute (GSSTI), Ghana,*
philusnarh@gmail.com

Rockefeller Rockefeller
*African Institute for Mathematical Sciences (AIMS), Limbe, Cameroon,*
rockfeller@aims-senegal.org

Jecinta Mulongo
*African Institute for Mathematical Sciences (AIMS),* Limbe, Cameroon
mulongo.jecinta@aims-cameroon.org

Marco Andrea Garuti
*African Institute for Mathematical Sciences (AIMS),* Limbe, Cameroon
marco@aims-cameroon.org

*Abstract*—The instability of power generation from national grids has led industries (e.g., telecommunication) to rely on plant generators to run their businesses. However, these secondary generators create additional challenges such as fuel leakages in and out of the system and perturbations in the fuel level gauges. Consequently, telecommunication operators have been involved in a constant need for fuel to supply diesel generators. With the increase in fuel prices due to socio-economic factors, excessive fuel consumption and fuel pilferage become a problem, and this affects the smooth run of the network companies. In this work, we compared four machine learning algorithms (i.e. Gradient Boosting, Random Forest, Neural Network, and Lasso) to predict the amount of fuel consumed by a power generation plant. After evaluating the predictive accuracy of these models, the Gradient Boosting model out-perform the other three regressor models with the highest Nash efficiency value of 99.1 %.

## I    INTRODUCTION

Energy crisis occurs when most part of the country experience blackouts as electricity supply falls behind demand, resulting in destabilising the national grid. Persistent load shedding of electricity in Cameroon has made the network operators to switch to generator plants in order to manage their businesses. Meanwhile, this mode of operation is subjected to fuel leaking in or out of the system and mechanical inaccuracies in the fuel level gauges. These perturbations increase the demand for fuel consumption. This work addresses the issue using machine learning techniques to measure two primary effects: (a) determine the feature parameters that influence fuel consumption and (b) generate a regression model to estimate the feature parameters in (a). The first research we conducted was to detect anomaly in power generation plants using supervised machine learning classification schemes. For more details, see Mulongo et al. [1].

This paper is organized as follows: Section 2 performs a descriptive statistic to extract the relevant feature parameters in the dataset. The expected fuel consumed is then measured based on the selected variables. Section 3 discusses the mathematics behind the four model regressors used in this work. Section 4 presents the model performance, and results are discussed in Section 5. Section 6 concludes the work.

## II    DATA DESCRIPTION

The dataset used for this work describes the routine maintenance of different base stations of the TeleInfra[1] network operator. The technicians manually record this data and the rate at which they collect the data depends on the period of maintenance. This work uses the data stream for one year collected with 31 features for 6109 observations. The Extra-tree applies to the data to extract the relevant features from the recorded data as shown in Fig. 1; we note that the feature, "fuel_per_period" is the most important feature followed by running time. This approach rejects the concept of adopting bootstrap copies of the training sample and instead of computing a maximum threshold for each one of the randomly chosen features at each node, it rather selects a threshold at random [2]. Note that this increases the accuracy of the selected features as well as reduces the effect of high variance and variance imbalance. The following features are directly related to the fuel consumption:

- The running time is the number of hours the generator was operating between two successive visits.
- The rate of consumption (in liter per hour, $L.h^{-1}$) refers to the quantity consumed by the generator in an hour and this is specific to each generator.
- The fuel per period is the quantity of fuel consumed by the generator between two successive visits.
- The consumption HIS stands for the fuel consumption of the BS per liter.

- The generator capacity (in kilo volt amperes KVA).
- The number of days

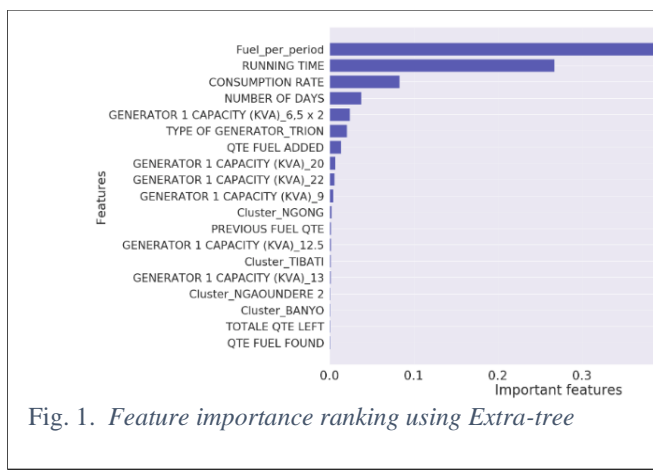

Fig. 1. *Feature importance ranking using Extra-tree*

Fig. 2 shows the distribution of the fuel used per period. The right-panel shows some outliers below 200 liters and above 800 liters, therefore, the dataset is slightly skewed with a skew of 0.49 and a kurtosis of -0.22 (the left-panel). The negative kurtosis means that the outlier in the data is less severe. We recorded a coefficient of determination (right-panel) of $R^2 \approx 97.60\%$. In addition, the rate of consumption has a direct influence on the fuel consumed as displayed in Fig. 3 (left-panel); this is confirmed in Fig. 3 (right-panel) where the running time has a strong impact with the fuel consumption.

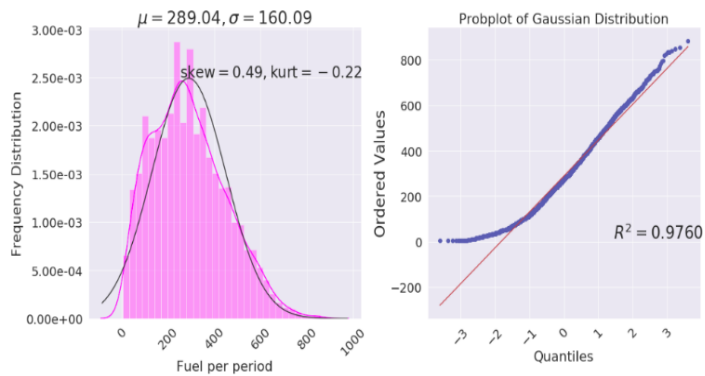

*Fig. 2. The distribution of fuel consumption. Left-panel is the histogram showing the fuel consumed is slightly positively skewed and the right-panel is the probability showing that the fuel consumed is ~ 97.60% Gaussian.*

### III METHODS

Below, we briefly describe the four regressors used in this work.

The first one is gradient boosting (GB). It is a type of boosting method – where errors are reduced sequentially, by combining several weak predictors while optimizing their output in order to provide a more relevant predictor. In this work, we use the GradientBoostingRegressor package in scikit-learn to build the additive model in a forward stage-wise way as described in [3].

The second regressor is random forest (RF) is an ensemble of decision trees where randomness is introduced into the learning process of each individual tree through a bagging[2] method as presented in [4]. Random forest is a sort of modification of bagging; a technique that focuses on the minimization of the estimated prediction variance [5]. Random forest works by building and averaging multiple decision trees during the training phase to reduce over-fitting and variance imbalance.

The third regressor used in our analysis is a neural network which is composed of many small units called neurons. These neurons are grouped into several layers such that the neurons of one-layer feed into the next layer through weight connections. A neuron takes an input value and then multiplies it by the connecting weight. The sum of all connected neurons is the put into an activation function which introduces a non-linearity in the network.

The fourth regressor used is the least absolute shrinkage and selection operator (Lasso). It is a regression analysis method that performs both variable selection and regularization in order to improve the prediction accuracy and interpretability of the model it produces [6]. The Lasso has some benefits such as helping to increase the model interpretability and to reduce over-fitting by removing irrelevant attributes that do not contribute in explaining the response variable This regressor can provide good performance prediction even in cases where we have few number of observations and a many number of features [7].

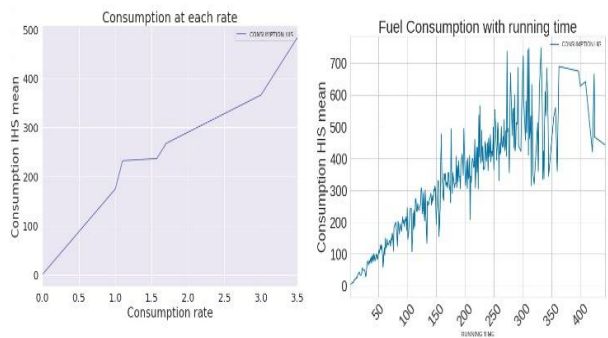

*Fig. 3. Left-panel: fuel consumption increasing with the increased of the rate of consumption. Right-panel: fuel consumption increasing with the increased of the running time.*

### IV PERFORMANCE EVALUATION

#### A Cross-Validation

Cross-validation is a statistical strategy of evaluating and examining in contrast learning algorithms by splitting data into two portions: the training set used to train the model and the validation set for validation. The major aim of this method is to estimate the performance of several learned models from the data while comparing these models in terms of their performances. In [8] many methods are presented to achieve this aim, namely, Hold-out validation, K-fold validation,

Leave-one-out cross-validation, and Repeated K-fold cross-validation. With the Hold-out validation, the data is split into two non-overlapping sub-data. The training set is used during the learning phase then the test set to estimate the error rate of the trained model. Note, this validation method uses only a portion of the data and the results strongly rely on a single training and/or test split. The K-Fold Cross validation works by partitioning the dataset into K number of fair subsets (folds) which k-1 are trained and tested with the K$^{th}$ dataset. In doing this, we end up with K different performance scores that can be summarized to get the mean and the standard deviation. The Leave-one-out cross-validation method is a specific case of the K-fold cross validation, where K is exactly the total number of observations in the dataset. Thus, at each iteration, the whole dataset except one observation is used to train the model, and a single held observation is used for the test. This estimation is unbiased due to its high variance and therefore, it is unreliable to use. Repeated K-fold cross validation is made up of reiterating the process of K-fold cross validation for a certain number of times. This helps to reduce high variance and therefore, reliable for this study.

### B  Evaluation of Predictive Accuracy

To evaluate how best our models use regressors to explain the response variable in order to figure out the most appropriate for the fuel consumption prediction, we need to analyze the predictive accuracy of each of these models that is, GB, RF ANN and Lasso. In this work, we used the Nash-Sutcliffe efficiency (NSE) and the error statistics to examine the predictive accuracy of these models.
The NSE is given as follows:

$$NSE = 1 - \frac{\sum_{k}^{N}(EST_k - OBS_k)^2}{\sum_{k=1}^{N}(OBS_k - \overline{OBS})^2}, \quad (1)$$

where $EST_k$ and $OBS_k$ refer to the k$^{th}$ estimated and observed fuel consumption values, respectively while $\overline{OBS}$ stands for the mean. As the NSE value gets close to one, the efficiency of the model improves. We computed four statistically errors, namely, Bias, Mean Absolute Error (MAE), Root Mean Square Error (RMSE) and the ratio of the RMSE and standard deviation (RSR). These metrics are very relevant, since they show error in the units or squared units of the target, which is useful in the analysis. Each of these metrics is defined as follows:

$$Bais = \frac{1}{N}\sum_{k=1}^{N}(OBS_k - EST_k) \quad (2)$$

$$MAE = \frac{1}{N}\sum_{k=1}^{N}|OBS_k - EST_k| \quad (3)$$

$$RMSE = \sqrt{\frac{1}{N}\sum_{k=1}^{N}(OBS_k - EST_k)^2} \quad (4)$$

$$RSR = \frac{\sqrt{\sum_{k=1}^{N}(OBS_k - EST_k)^2}}{\sqrt{\sum_{k=1}^{N}(OBS_k - \overline{OBS})^2}} \quad (5)$$

### C  RESULTS AND DISCUSSION

The residual reported in Fig. 4 shows how the observed error (residuals) is consistent with the predicted error; the residuals are well estimated about 0 with both recording a test score of $R^2 \approx 99\%$ as compared to the residual values for multilayer perceptron (MLP) and Lasso regressors. These residuals are produced by computing the differences between the observed values of the target variable and the predicted values displaying the portions within the target that are prone to more or less error

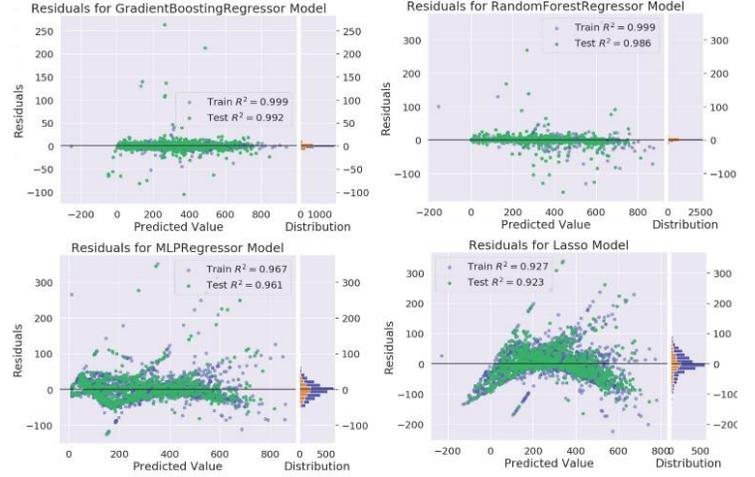

Fig. 4. *Residual displaying the variance of the the errors of the selected regressors.*

Fig. 5 (top panel) confirms how the GB regressor accurately predicts the patterns of the observed values. The right-panel of this figure shows an accuracy score of $R^2 = 99.2\%$ for this regressor.

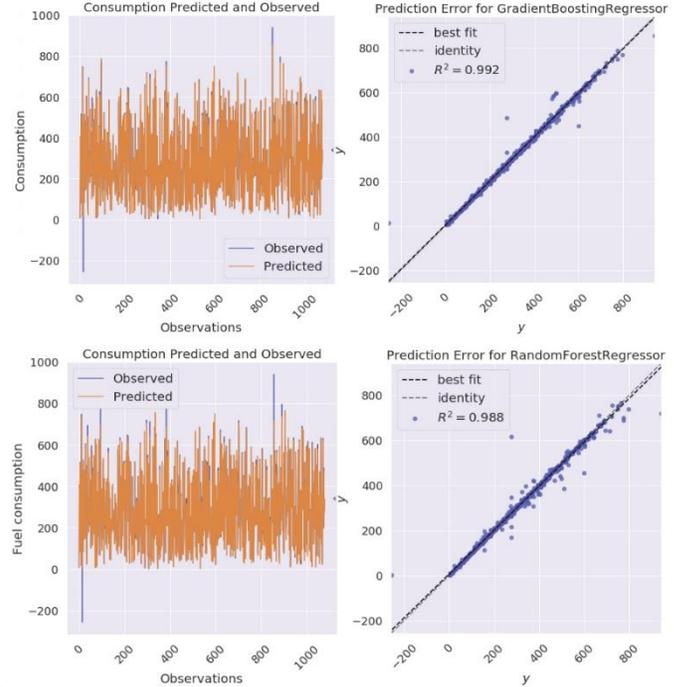

.
Fig. 5. Top-left: the GB regressor accuracy for both the values of the fuel consumed and estimated ones. Top-right: the error recorded for using the GB regressor. The fiducial marks (thick black dashes) define the frame of reference for the spatial measurement. Bottom-left: the RF regressor accuracy for both the real values of the fuel consumed and predicted ones. Bottom-right: the error recorded for using the RF regressor. The fiducial marks (thick black dashes) define the frame of reference for the spatial measurement.

The bottom-panel of Fig. 5 shows the RF regressor with the prediction precision of fuel consumption of $R^2$ = 98.8 % making it the second-best model to predict the fuel consumption. In Fig. 6 (top panel), the Lasso function estimated the fuel consumed at an accuracy of $R^2$ = 93.0 % %. Fig. 7 shows the MLP where the biases are initialized to unity. The overall error for the MLP is 0.279 making it a good network for predicting fuel consumption; this is supported by Fig. 6 (bottom panel) recording a precision score of $R^2$ = 96.1%.

The 10-folds cross-validation was used to estimate the performance of the learning model from the data. Also, we fit the learning curve to show the relationship between the training and cross validation test score. In Fig. 8 (bottom-left) the learning rate of MLP converges faster than the others, which need over 2500 training instances. The slow convergence of these regressors is due to the limited number of observations for the study.

errors as compared to other model regressors. Hence, making GB the best predictor for this work.

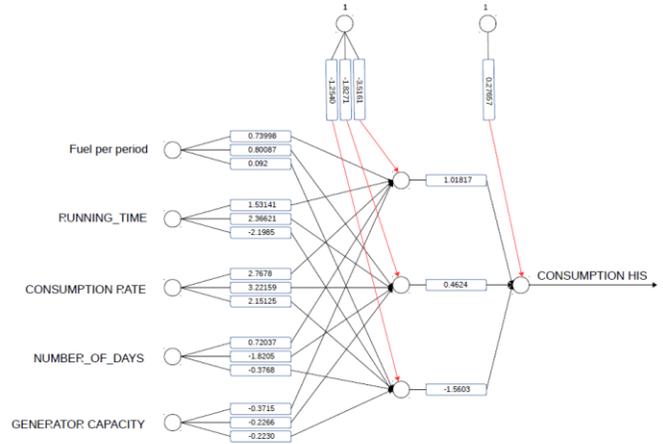

Fig. 7. *Neural network diagram showing the input variables used with their respective weights and one hidden layer.*

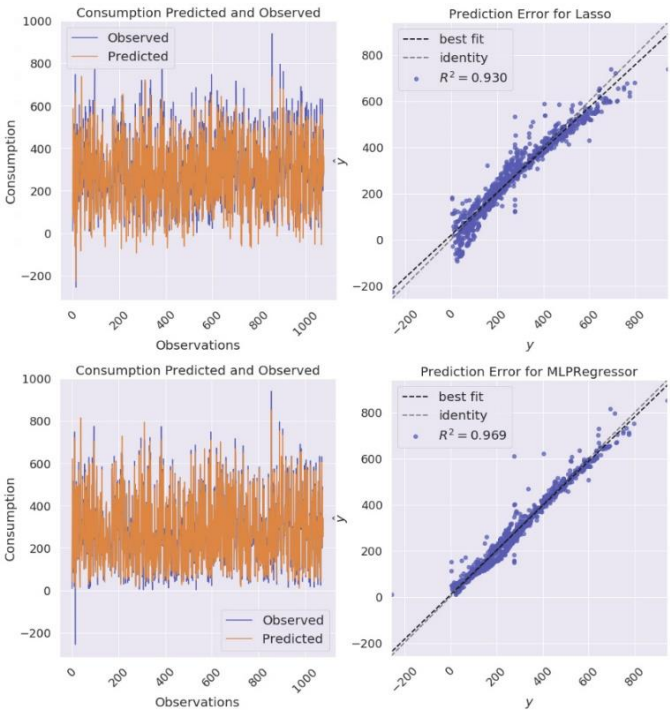

Fig. 6. *Top-left: Lasso regressor accuracy for both the real values of the fuel consumed and estimated ones. Top-right: the error recorded for using the Lasso regressor. The fiducial marks (thick black dashes) define the frame of reference for the spatial measurement. Bottom-left:MLP regressor accuracy for both the real values of the fuel consumed and validated ones. Bottom-right: The error recorded for using the MLP regressor. The fiducial marks (thick black dashes) define the frame of reference for the spatial measurement.*

The NSE values displayed in Table 1 are computed using the 10-folds cross-validation scores. Again, the GB regressor produced the highest score (99.1 %) followed by RF (98.6 %). In Table 2, the model GB recorded the least average prediction

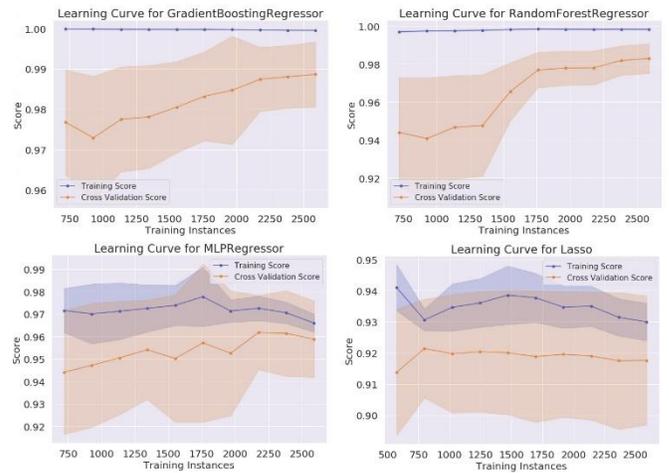

Fig. 8. *The distribution of 10-folds cross-validation between the observed values of fuel consumption and the regressor models.*

Table I. NSE results

| Model | Nash-Sutcliffe Efficiency |
|---|---|
| Gradient Boosting | 0.991 |
| Random Forest | 0.986 |
| Neural Network (MLP) | 0.961 |
| Lasso | 0.917 |

Table II. Error estimation model regressors.

| Error Statistic | Gradient Boosting | Random Forest | Neural Network | Lasso |
|---|---|---|---|---|
| BIAS | -0.11 | 0.95 | -0.08 | 0.10 |
| MAE | 4.68 | 5.30 | 20.40 | 32.41 |
| RMSE | 12.54 | 12.54 | 29.82 | 45.18 |

## V CONCLUSION

We have investigated four algorithms namely: GB, RF, ANN, and Lasso to build up four predictive models. Some of the key findings captured in this work are as follows:
rate of consumption,

- The relevant factors needed to predict fuel consumption are the running time of the generator, the fuel per period, the rate of consumption, the generator capacity, and the number of days.
- The residual presented in Fig. 4 shows that the GB regressor is susceptible to less error, with the distribution of the error being symmetric. Thus, the points are uniformly scattered around the horizontal axis, making this to have the best model performance ($R^2 \approx 99$ %) than the other models, followed by RF with a model performance $R^2 = 98$ %.
- The most relevant hyper-parameters to fine tune using the RF model are the number of trees and the size of the random sample of the features to consider at each separation. Tuning these parameters decreases the variance in of the ultimate model.
- With the GB model, the relevant hyper-parameters to fine tune are the number of trees, the learning rate, and the depth of the trees. Fine tuning these parameters improves the performance of the model accuracy.

### ACKNOWLEDGMENT

The author would like to thank TeleInfra Company and Group One Holding Company for providing the dataset for this study. T. Ansah-Narh is very grateful to the Data Science Intensive (DSI) Program, organized by the African Institute for Mathematical Sciences (AIMS) South Africa, in conjunction with SA-DISCNet.